\documentclass{article}

\usepackage{arxiv}

\usepackage[utf8]{inputenc} 
\usepackage[T1]{fontenc}    
\usepackage{hyperref}       
\usepackage{url}            
\usepackage{booktabs}       
\usepackage{amsfonts}       
\usepackage{nicefrac}       
\usepackage{microtype}      
\usepackage{lipsum}		
\usepackage{graphicx}
\usepackage{natbib}
\usepackage{doi}
\usepackage{algorithm}
\usepackage{algorithmic}
\usepackage{amssymb}
\usepackage{xcolor}         
\usepackage{subfigure}
\usepackage{amsmath}
\usepackage{amsthm}
\usepackage{multirow}
\usepackage{diagbox}
\usepackage{enumitem}
\usepackage{bm}
\usepackage{color, colortbl}

\definecolor{COLOR_MEAN}{HTML}{f0f0f0}

\title{Maybe Only 0.5\% Data is Needed: A Preliminary Exploration of Low Training Data Instruction Tuning}

\author{
    {Hao Chen \thanks{ \quad Equal contribution and shared co-first authorship.}} \\
    Zhejiang University \\
    \texttt{h.c.chen@zju.edu.cn} \\
    \And
    {Yiming Zhang \footnotemark[1]} \\
    Zhejiang University \\
    \texttt{yimingz@zju.edu.cn} \\
    \And
    {Qi Zhang \footnotemark[1]} \\
    Zhejiang University \\
    \texttt{cheung\_se@zju.edu.cn} \\
    \AND
    {Hantao Yang \footnotemark[1]} \\
    Zhejiang University \\
    \texttt{ht.yang@zju.edu.cn} \\
    \And
    {Xiaomeng Hu} \\
    Zhejiang University \\
    \AND
    {Xuetao Ma}\\
    ZhongHao XinYing (Hangzhou) \\
    Technology Co., Ltd. \\
    \texttt{maxuetao@gmail.com} \\
    \And
    {Yifan Yanggong} \\
    ZhongHao XinYing (Hangzhou)\\
    Technology Co., Ltd. \\
    \texttt{baihu@cltech.com}
    \And
    {Junbo Zhao \thanks{\quad Corresponding author.}} \\
     Zhejiang University \\
    \texttt{j.zhao@zju.edu.cn} \\
}

\renewcommand{\shorttitle}{\textit{arXiv} Template}

\hypersetup{
pdftitle={Maybe Only 0.5\% Data is Needed: A Preliminary Exploration of Low Training Data Instruction Tuning},
pdfsubject={cs.AI},
pdfkeywords={First keyword, Second keyword, More},
}

\begin{document}
\maketitle

\begin{abstract}
Instruction tuning for large language models (LLMs) has gained attention from researchers due to its ability to unlock the potential of LLMs in following instructions. While instruction tuning offers advantages for facilitating the adaptation of large language models (LLMs) to downstream tasks as a fine-tuning approach, training models with tens of millions or even billions of parameters on large amounts of data results in unaffordable computational costs. To address this, we focus on reducing the data used in LLM instruction tuning to decrease training costs and improve data efficiency, dubbed as Low Training Data Instruction Tuning (LTD Instruction Tuning). Specifically, this paper conducts a preliminary exploration into reducing the data used in LLM training and identifies several observations regarding task specialization for LLM training, such as the optimization of performance for a specific task, the number of instruction types required for instruction tuning, and the amount of data required for task-specific models. The results suggest that task-specific models can be trained using less than \textbf{0.5\%} of the original dataset, with a \textbf{2\%} improvement in performance over those trained on full task-related data.

\end{abstract}


\section{Introduction}
\label{sec:intro}
\begin{figure}[ht]
\begin{center}
   \includegraphics[width=0.7\linewidth]{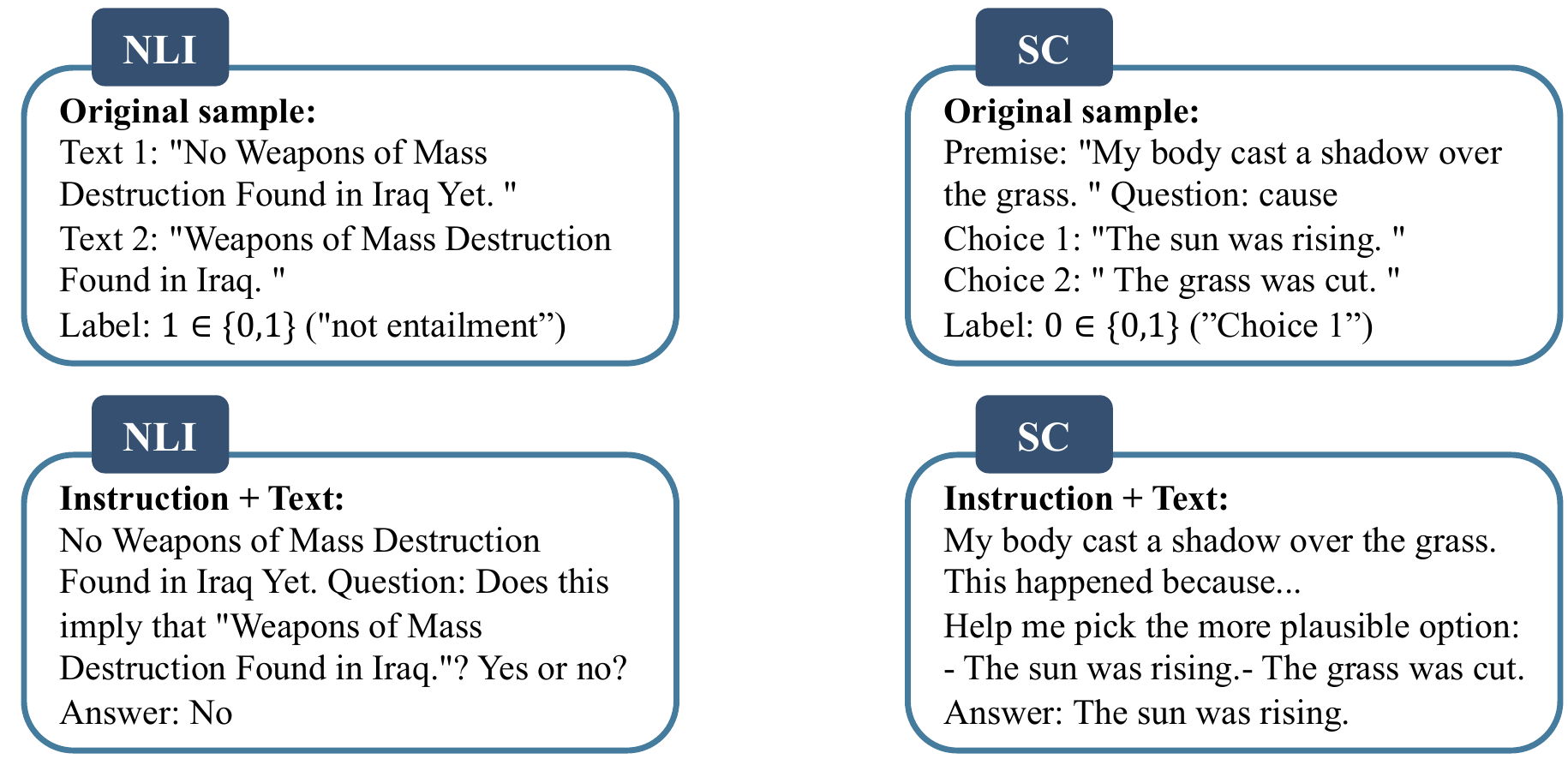}
\end{center}
   \caption{An illustration of the different between fine-tuning and instruction tuning, taking the task of natural language inference and causal reasoning as examples. LLMs predict labels for samples in fine-tuning while answering questions for the instruction set in instruction tuning. The Natural Language Inference(NLI) task involves determining the logical relationship between two pieces of text, typically referred to as the "premise" and the "hypothesis." The goal of NLI is to determine whether the hypothesis is true, false, or undetermined based on the information provided in the premise. Sentence Completion(SC) involves predicting the most likely word or sequence of words to complete a given sentence or phrase.}
\label{fig:intro}
\end{figure}

With the tremendous momentum of large language models and their impressive performance~\citep{gpt4, taylor2022galactica, touvron2023llama, zhao2023survey}, instruction tuning as one of their adaptation tuning approaches with fine-tuning on samples described via instructions~\citep{flan, chung2022scaling, wei2022finetuned}, has attracted much attention from researchers~\citep{instructGPT}. Instruction tuning means fine-tuning the large language models on samples described via instructions ~\citep{flan}, as shown in Fig.~\ref{fig:intro}. Yet lacking comprehensive analysis, this newly presented tuning method has shown its superior power in unlocking the endowed abilities of LLMs of the instruction following. Among related works, instruction tuning is mainly used to align LLMs with humans~\citep{instructGPT}, generalize to certain unseen tasks~\citep{T0P3, wei2022finetuned}, and specialize to a certain downstream task~\citep{self-instruct, elm}. Compared to conventional fine-tuning, instruction tuning offers the advantages of requiring fewer data and being more human-friendly, and many researchers also regard instruction tuning as a new fine-tuning approach to adapt LLMs for corresponding downstream tasks~\citep{wei2022finetuned, puri2022many, elm}.

Although instruction-tuned LLMs are very powerful, training models with tens of millions or even billions of parameters often face the problem of training costs and, according to scaling laws~\citep{kaplan2020scaling}, require large amounts of data. Most current works related to instruction tuning always expand the amount of data or the variance of instructions used for instruction tuning, for instance, the FLAN~\citep{flan} collection containing 15M examples covering 62 tasks, and the P3 dataset~\citep{T0P3} containing 250B tokens with 10 instruction types. However, with the increment of the LLMs' scale from 400M~\citep{devlin2019bert} to 540B~\citep{chowdhery2022palm} or even larger scale in the future, the scale of the training data used for instruction tuning will greatly affect the training costs. For example, OpenAI has listed the costs for instruction tuning the Davinci~\citep{openaipricing} with 0.03 dollars per 1k token, which means training a model with P3 will cost 7.5M dollars.

However, we argue that there is a lack of instruction-tuning-related research on reducing the amount of data used for the training stage to decrease the training costs.
While training costs are often related to hardware conditions and engineering operations, data efficiency can be improved with the help of algorithms, thus reducing the cost of training data. For example, self-instruct~\citep{self-instruct} declines the number of instances corresponding to each instruction and uses 80K examples to instruct-tune the Davinci with 338 dollars cost. We may call these methods reducing the training data scale during the instruction tuning as Low Training Data Instruction Tuning (LTD Instruction tuning) in the below.

Considering instruction tuning as a fine-tuning method for specialization tasks, we explore another line of LTD instruction tuning -- reducing the diversity of tasks and instructions. While most instruction tuning-based works focus on the generalization ability of LLMs, few focus on specialization. What if the LLM only needs to be tuned for a certain task? How many instructions do we need? How many training examples are needed by the model? These questions are still under-explored.

In this paper, we conduct a preliminary exploration into reducing the data used in the LLM training phase, from the perspective of the data itself, to decrease the training costs and improve data efficiency. Specifically, we aim to identify the most valuable core samples from existing data to help the model acquire the knowledge necessary for downstream tasks, and to achieve comparable or even better performance with only a small amount of data. As a result, after selecting a specific task and instruction format, we successfully train a task-specific model using less than \textbf{0.5\%} of the original dataset, with a comparable performance compared to the model trained on task-related data in P3. Our observations on the natural language inference (NLI) task are as follows, and they yield several key findings regarding task specialization for LLM training, which we hope can provide some insights for the community.

\begin{itemize}
    \item If only to optimize performance for a specific task, an LLM model tuned solely on target task data is likely to outperform a modal tuned on data from different types of tasks.
    \item When specializing in a single task, it appears that only one instruction may be sufficient for instruction tuning. While increasing the number of instruction types can improve performance, the marginal effect becomes less significant, and there may even be cases where a single instruction outperforms ten types of instruction.
    \item In contrast to training a model for overall task performance, our results also suggest that 16k instances (1.9M tokens, 0.5\% of the P3) may be sufficient to train an NLI task-specific model.
\end{itemize}

\section{Related Work}
\label{sec:relatedwork}
\paragraph{Large language models (LLMs).}
Large language models (LLMs) generally refer to language models with tens or hundreds of billions of parameters and are trained with massive data, e.g., GPT-3~\citep{gpt3}, GPT-4~\citep{gpt4}, Galactica~\citep{taylor2022galactica}, LLaMA~\citep{touvron2023llama}, etc. Many studies have shown such models have under-explored emergent abilities compared to small models when the scale exceeds a certain level~\citep{kaplan2020scaling, wei2022emergent}. \cite{zhao2023survey} concludes that the current emergent abilities of large models include mainly in-context learning, which helps models possess the ability to adapt to the downstream tasks without gradient update, but only a few examples or several task demonstrations~\citep{ye2023context, dong2023survey}. Some researchers also regard this operation as instruction~\citep{ye2023context}. Another emergent ability is instruction following~\citep{instructGPT}. By adding task descriptions to the data, the LLMs can understand the requirements of a task without additional samples on unseen downstream tasks~\citep{promptengineer, si2022prompting, wei2022finetuned, sumers2022talk}, or tuning with these re-formatted data to endow task-specific capability to the model~\citep{flan, self-instruct, T0P3, gupta2022improving, chung2022scaling, DEFT, elm}. Many works also find an important ability of step-by-step reasoning (also known as chain-of-thoughts)~\citep{COTpaper, wang2023selfconsistency}, which helps LLMs to derive a final answer by splitting the task and using a form of intermediate steps of reasoning.

\paragraph{Instruction tuning.}
Although current instruction-following LLMs have demonstrated strong performance in deriving task-relevant answers relying on instruction~\citep{si2022prompting, wei2022finetuned, T0P3, promptengineer}, when facing task-specific issues, fine-tuning is still the preferred option for achieving better results~\citep{i-learning_survey, zhao2023survey}. Unlike only using instructions without training to guide the model output, instruction tuning is an approach to fine-tune LLMs with data fused with instruction to achieve task-specific effects~\citep{T0P3, flan, puri2022many, elm, DEFT, Hint}. Most works focus on the generalization capability brought by instruction tuning, which helps the model to have cross-task generalization by fine-tuning on multi-task instruction data~\citep{T0P3, flan}. In addition, instruction tuning can also help the model improve performance on a specific task~\citep{self-instruct, elm, DEFT, Hint}, and research has shown that on single-task fine-tuning, instruction tuning can accelerate model convergence. 

\paragraph{Low training data.}
While the effectiveness of LLMs is superb, the training costs associated with huge parametric models have simultaneously limited the popularity and adoption of LLMs~\citep{hoffmann2022training, zhao2023survey}. Many works try to explore the possible cost reduction in LLMs from the perspective of data. \cite{elm} report an expert LLM fine-tuned on one single task can outperform a multi-task tuned LLM. Self-instruct~\citep{self-instruct} starts from the instruction generation, which means a model generates its own instructions or prompts and learns to follow them. DEFT~\citep{DEFT} assumes the presence of unlabeled task-related data, and by retrieving data using K-nearest neighbors from the data pool that is highly similar to the task data, new fine-tuned models can achieve the same performance as full dataset-trained models do. HINT~\citep{Hint} incorporates instruction into model parameters to reduce the number of tokens corresponding to instruction in each input training to save the token cost.

\section{Method}
As mentioned in the introduction, our key ideas for limiting the data scale are reducing the variety of instructions and focusing on specialized tasks. 
And in this section, we'll introduce how to reduce the entire dataset scale based on both ideas separately.

\subsection{Motivation}
We start by introducing the motivation for our method before explaining its details. At present, the development of LLM has received tremendous attention, but the high training cost brought by the model with large parameters also limits the popularity and application of LLM. We hope to explore how to improve the efficiency of LLM from the perspective of reducing its training cost. The main costs in the current use of LLM include training costs and data costs~\citep{zhao2023survey}. The training costs include using public APIs or self-finetuning large models, which mainly face hardware or paralleling requirements, and the algorithm plays a light role in it, while for data costs, data-centric algorithms can be used. Therefore, we hope to explore how to reduce the data used in LLM training from the perspective of the data itself to reduce training costs and improve the efficiency of data usage. That is, to retrieve the most useful core samples from the existing data to help the model learn the knowledge required for downstream tasks, and to achieve good performance with only a small amount of samples.

\begin{figure}[ht]
\begin{center}
   \includegraphics[width=0.8\linewidth]{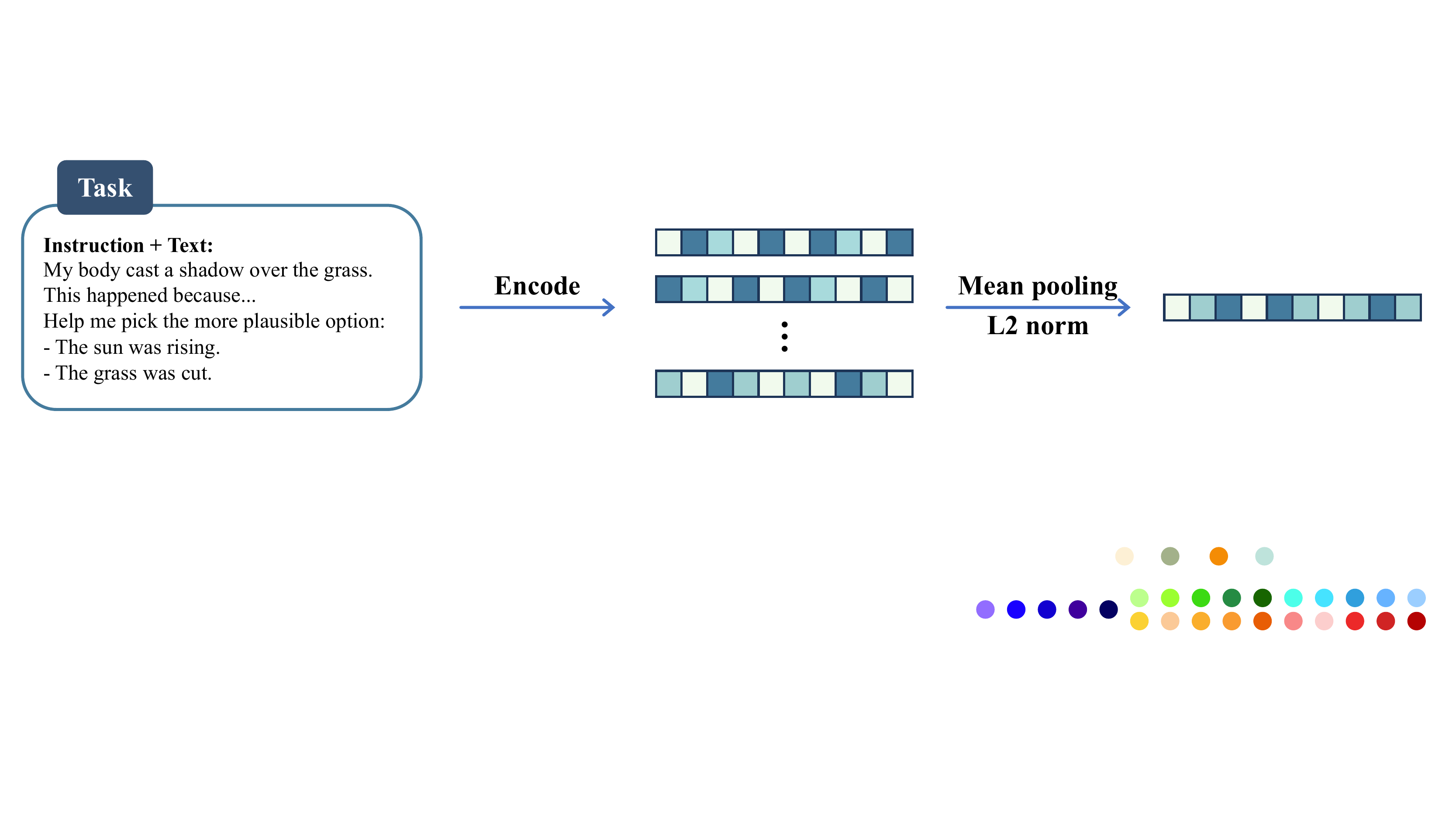}
\end{center}
   \caption{The pipeline of our proposed method. First, encode each sentence into embeddings and pre-process with mean pooling and L2 normalization. After this, in the latent space, we cluster all sample points into a few clusters. Then employ the sampling methods on these clustered samples to find the core samples of the original distribution. Finally, using these retrieved samples to instruction tuning the LLMs and make the evaluation. The three rectangles represent the latent space, and one color series in the latent space refers to one task. Points of the same color series but different shades correspond to data from the same task but from different datasets, e.g., the NLI task has five datasets, thus making it five different shades.}
\label{fig:method}
\end{figure}

\subsection{Coreset-based Selection}
Due to the vague boundaries of NLP tasks~\citep{zhao2023survey}, samples from different tasks may have low discriminability, and it is often infeasible to manually select more suitable samples for NLP tasks. Therefore, we propose a coreset-based task-related data selection method to automatically retrieve core samples from the task dataset, to help train task-specific LLM with fewer samples. Specifically, this algorithm includes the following steps:

\paragraph{Sentence embedding and pre-processing} 
Firstly, we re-formatted the data to the training input format used during the instruction tuning training phase, i.e., data with description instructions, adding answers at the end to format one complete training data. Then, we encode all samples using a pre-trained language model (e.g., Galactica~\citep{taylor2022galactica} or Bert~\citep{devlin2019bert}). Specifically, we extract the \textit{last\_hidden\_state} of each sample after feeding the model as the word embeddings or each sentence. Note that masked language models like Bert~\citep{devlin2019bert} have \textit{cls} token as the sentence embedding for one input sentence, but for generative models like GPT series~\citep{gpt3, gpt4}, they do not have this token. Following ~\cite{reimers2019sentencebert}, we performed mean pooling on the word embeddings of each sample and obtained a one-dimensional vector as the sentence embedding for this sample. To speed up the computation and facilitate vector similarity calculation, we normalize all sentence embeddings to length 1, i.e., L2 normalization is performed on the embedding dimension.

\paragraph{Clustering}
In the clustering step, we take into account that the fuzziness of NLP task boundaries may cause little variation among samples from different tasks. Thus we approach unsupervised clustering by focusing on data representations, rather than relying on label information to group data points together based on the same categories or tasks. Specifically, after obtaining the sentence embeddings from the first step, we perform unsupervised clustering using K-Means~\citep{lloyd1982least} in the embedding space to obtain the mapping of each sample and its corresponding cluster label. Then, based on the frequency of samples from one downstream task appearing in several clusters, we select the center point of the cluster with the highest frequency as the distribution center point of that downstream task. Next, for all the samples in the task, we calculate the cosine similarity with the distribution center point (the choice of distance function has little effect on the results, and we follow \cite{gpt4} to choose cosine similarity), and find the closest sample from task data to this center point as the task center point. Note that the distribution center point is the center of this task data in embedding space, which may not exist in the task data, while the task center point is one exact sample from this task data with the biggest cosine similarity to the distribution center point.

\paragraph{Coreset sampling}
Intuitively, after obtaining the distribution center point corresponding to the downstream task, we can select the most similar sample as the representative task sample based on cosine similarity, as done in~\citep{DEFT}, which achieved good results. However, their retrieval method selects high-similarity samples from the data pool based on existing samples to improve task performance, which can be considered as a form of data augmentation through retrieval. This contradicts our goal of reducing the data required for training. We aim to find a small set that approximates the distribution of the full dataset using as few samples as possible. Therefore, the K-nearest neighbor method in DEFT~\citep{DEFT} is unsuitable for this situation since samples with high similarity do not approximate the full set distribution~\citep{sener2018active}. To achieve the sampling of core samples, we used one coreset algorithm KCentergreedy~\citep{sener2018active}, which aims to choose $k$ center points such that minimize the largest distance between a random data point and its nearest center, and has been proven efficient in obtaining a set of core samples of one distribution. 

We use the task sample center point as the initial center, feed all the sentence embeddings of task samples obtained in the previous steps, and use the KCenterGreedy algorithm to collect a set of core samples from the task samples according to the given proportion. The subset of the original task dataset collected can achieve the same or even higher performance with fewer data.

\subsection{To Be continue...}
It should be noted that in addition to this method, we also explored two other ways to reduce the training data required by fine training.
However, due to computing power and time limitation, they are not yet complete enough to be reported.
Please wait for our future papers.

\section{Experiments}

\subsection{Setup}
\paragraph{Dataset}
Following the setup in P3\citep{T0P3}, we conduct experiments on a total of 11 datasets, which spanned across 4 NLP tasks, namely Natural Language Inference (NLI, 1.9M tokens), Sentence Complement (SC, 660.6K tokens), Word Sense Disambiguation (WSD, 25.5K tokens) and Coreference Resolution (CR, 185.1K tokens). On the contrary, the full task-related dataset of P3 contains 382.8M tokens.

To be specific, for the Natural Language Inference task, we employ RTE \citep{dagan2006pascal}, CB \citep{wang2020superglue}, and ANLI \citep{nie2020adversarial} datasets, while for the Sentence Complement task, we used COPA \citep{wang2020superglue}, HellaSwag~\citep{zellers2019hellaswag}, and Story Cloze~\citep{mostafazadeh2016corpus}. For the Coreference Resolution task, we utilized Winogrande \citep{sakaguchi2019winogrande} and WSC \citep{wang2020superglue} datasets, and for the Word Sense Disambiguation, we used WIC \citep{wang2020superglue}. Moreover, to generate the instruction-style dataset, we randomly selected only one prompt from each dataset.

\paragraph{Model}
We utilize the Galactica-1.3b \citep{taylor2022galactica} model to conduct experiments in our study. Galactica models are trained on a vast scientific corpus and are tailored to handle scientific tasks such as citation prediction, scientific question-answering, mathematical reasoning, summarization, document generation, molecular property prediction, and entity extraction. Following~\citep{gpt3}, similar to the pre-training phase, we treat all datasets as next token prediction tasks. In particular, we employ the AdamW optimizer with a learning rate of 1e-5.

\paragraph{Evaluation}
Prior research on instruction tuning failed to state the evaluation methods utilized explicitly. In this paper, we introduce our evaluation methodology, which can serve as a reference for other researchers working in this area.

When a tokenized sequence $x$ and a tokenized answer option $y$ (with a length of $l$) are provided as input to the model, a probability matrix $P_{l\times vocab\_size}$ is generated. Subsequently, for an answer option $y^i$, its corresponding probability $p^i$ can be obtained by multiplying the probabilities of each token in $y^i$ using the formula $\prod^{l^i}_{j=1} p^i_j$. The answer option with the highest probability is considered by the model as the optimal answer.

\subsection{Results on Natural Language Inference Tasks}
We describe in this section the NLI task results of using our method, as seen in Table.~\ref{tab:nli_results}. According to the information presented in this table, when considering a specific task (NLI in this case), our method achieves a performance improvement of \textbf{2\%} on average beyond the baseline (P3 in table) on the NLI task, using only \textbf{0.5\%} of the available data from P3. 
In comparison to using all ten instructions from P3, we find that selecting only one instruction allows us to achieve comparable results to using the entire dataset from P3 with only 10\% of the data. 

Regarding task-specific models, the second and fourth rows have shown that the diversity of tasks might have a negative impact. 
Moreover, by exclusively utilizing data from the NLI task, we obtain results that are approximately 8\% on average higher than those from P3. 

Therefore, we speculate that for task-specific requirements, using only relevant data for the target task and a single instruction may be more effective than directly employing large-scale models. 
Notice that these observations may only be applicable to the NLI task, as other tasks remain largely unexplored due to computational limitations.

\begin{table}[H]
\centering
\begin{tabular}{c|cccccc}
\toprule[1.2pt]
\textbf{Model} & RTE & CB & ANLI R1 & ANLI R2 & ANLI R3 & Avg.  \\ \midrule
Vanilla Model (0\%) & 54.51 & 41.07 & 33.40 & 33.40 & 33.58 & 39.19 \\ 
P3 (100\%) & 76.17 & 75.00 & 44.00 & 35.70 & 39.42 & 54.06 \\ 
Fixed Instruction (10\%) & 71.11 & 66.07 & 43.60 & 38.90 & 42.17 & 52.37 \\ 
NLI-related (5\%) & 79.06 & 82.14 & 60.40 & 46.50 & 46.67 & 62.95 \\ 
NLI coreset (0.5\%) & 74.73 & 73.21 & 49.60 & 41.90 & 43.75 & 56.64 \\ \bottomrule[1.2pt]
\end{tabular}%
\setlength{\abovecaptionskip}{5pt}%
\caption{Test accuracy (\%) with used models on NLI task. The first row suggests the performance of the vanilla model (Galactica-1.3b here) without instruction tuning. P3 stands for using a full task-related dataset from P3 (10 instructions with 11 datasets). Fixed instruction stands for using only one instruction type, resulting in a 10\% of P3. Among all 11 datasets, NLI task-related data accounts for ~50\%, thus NLI-related refers to training with only 5\% of P3 (NLI data). With this, NLI coreset indicates using only 10\% of NLI task-related data, thus making a 0.5\% of P3.}
\label{tab:nli_results}
\end{table}

\subsection{Ablation Study for Sampling}
Regarding the strategy for model sampling, we have also explored several alternative sampling approaches. 
The table indicates that selecting the most similar or dissimilar data based on cosine similarity yields significantly poor results, even lower than the vanilla model. 
We speculate that this outcome may be attributed to an imbalance issue when sampling solely based on similarity within the task pool consisting of these five datasets. 
It is likely that a majority of the sampled examples are dominated by RTE or ANLI data.

\begin{table}[H]
\centering
\begin{tabular}{c|cccccc}
\toprule[1.2pt]
\textbf{Sampling Method} & RTE & CB & ANLI R1 & ANLI R2 & ANLI R3 & Avg.  \\ \midrule
Vanilla Model (0\%) & 54.51 & 41.07 & 33.40 & 33.40 & 33.58 & 39.19 \\ 
coreset (0.5\%) & 74.73 & 73.21 & 49.60 & 41.90 & 43.75 & 56.64 \\ 
topK (0.5\%) & 47.29 & 10.71 & 33.40 & 33.20 & 33.92 & 31.70 \\ 
leastK (0.5\%) & 50.18 & 8.93 & 33.80 & 33.60 & 34.08 & 32.12 \\ 
mixed (0.5\%) & 47.29 & 8.93 & 33.30 & 33.30 & 33.50 & 31.26 \\ \bottomrule[1.2pt]
\end{tabular}%
\setlength{\abovecaptionskip}{5pt}%
\caption{Ablation study on test accuracy (\%) of NLI task using 10\% data (0.5\% of full task-related dataset from P3). The first row represents the vanilla model without any sampling method (0\% of P3). Coreset refers to our method of using core samples of the NLI dataset. TopK uses samples that are close to the NLI distribution center point, while the leastK uses the least close samples. Mixed indicates mixing top and least close samples. The number of samples used is the same for all methods (10\% of NLI task, \textasciitilde16k samples).} 
\label{tab:ablation}
\end{table}

\section{Conclusion and Future Work}
In this paper, we present experimental results from our preliminary exploration of Low Training Data Instruction Tuning, by tuning the Galactica-1.3b Model on the P3 dataset for the NLI task. Our study has revealed several findings: 
\begin{enumerate}
    \item task-specific models may benefit from fixed task types to achieve superior performance;
    \item the diversity of instruction formats may have minimal impact on task-specific model performance;
    \item even a small amount of data (1.9M tokens) can lead to promising results in instruction tuning for task-specific models.
\end{enumerate} 
It should be noted that our work has several limitations due to the constraints of computational resources, such as conducting experiments solely on Galactica-1.3b and utilizing only the NLI task data from the P3 dataset. 

We hope our preliminary findings can provide insights to the community on Low Training Data Instruction Tuning, and yield a new perspective on instruction tuning for researchers.
As for future work, we plan to validate these ideas on bigger models using a more comprehensive range of tasks and datasets.

\bibliographystyle{unsrtnat}
\bibliography{references} 

\end{document}